\pdfoutput=1

\documentclass[11pt]{article}

\usepackage[final]{acl}
\usepackage{booktabs}

\usepackage{times}
\usepackage{latexsym}
\usepackage{amsmath} 
\usepackage[multiple]{footmisc}
\usepackage[T1]{fontenc}

\usepackage[utf8]{inputenc}

\usepackage{microtype}

\usepackage{inconsolata}
\usepackage{hyperref}

\usepackage{graphicx}

\usepackage{tabularx}
\newcolumntype{b}{X}
\newcolumntype{s}{>{\hsize=.5\hsize}X}

\usepackage{tikz}
\usepackage{pgfplots}
\pgfplotsset{compat=newest}
\usepgfplotslibrary{colorbrewer}
\usepgfplotslibrary{statistics}
\usetikzlibrary{patterns}
\usepackage{booktabs}
\newcommand{\mysubsection}[1]{\vspace{0.3em}\noindent\textbf{#1}}

\title{Do Multilingual Large Language Models Mitigate Stereotype Bias?}

\author{
  \textbf{Shangrui Nie\textsuperscript{1,3}},
  \textbf{Michael Fromm\textsuperscript{2,3}},
  \textbf{Charles Welch\textsuperscript{1,3}},
  \textbf{Rebekka Görge\textsuperscript{2,3}},
  \textbf{Akbar Karimi\textsuperscript{1,3}},\\
  \textbf{Joan Plepi\textsuperscript{1,3}},
  \textbf{Nazia Afsan Mowmita\textsuperscript{2,3}},
  \textbf{Nicolas Flores-Herr\textsuperscript{2,3}},
\textbf{Mehdi Ali\textsuperscript{2,3}},
    \textbf{Lucie Flek\textsuperscript{1,3}}
\\
\\
  \textsuperscript{1}Conversational AI and Social Analytics (CAISA) Lab, University of Bonn, Germany
\\  
  \textsuperscript{2} Fraunhofer Institute for Intelligent Analysis and Information Systems (IAIS), Germany
\\
  \textsuperscript{3} Lamarr Institute
for Machine Learning and Artificial Intelligence, Germany\\
  \small{
       \texttt{\url{http://lamarr-institute.org/research/natural-language-processing/} }
}
}

\begin{document}
\maketitle
\begin{abstract}
While preliminary findings indicate that multilingual LLMs exhibit reduced bias compared to monolingual ones, a comprehensive understanding of the effect of multilingual training on bias mitigation, is lacking. This study addresses this gap by systematically training six LLMs of identical size (2.6B parameters) and architecture: five monolingual models (English, German, French, Italian, and Spanish) and one multilingual model trained on an equal distribution of data across these languages, all using publicly available data. To ensure robust evaluation, standard bias benchmarks were automatically translated into the five target languages and verified for both translation quality and bias preservation by human annotators. Our results consistently demonstrate that multilingual training effectively mitigates bias. Moreover, we observe that multilingual models achieve not only lower bias but also superior prediction accuracy when compared to monolingual models with the same amount of training data, model architecture, and size. 


\end{abstract}

\section{Introduction}
\begin{figure}[t]
    \centering
    \includegraphics[scale=0.58]{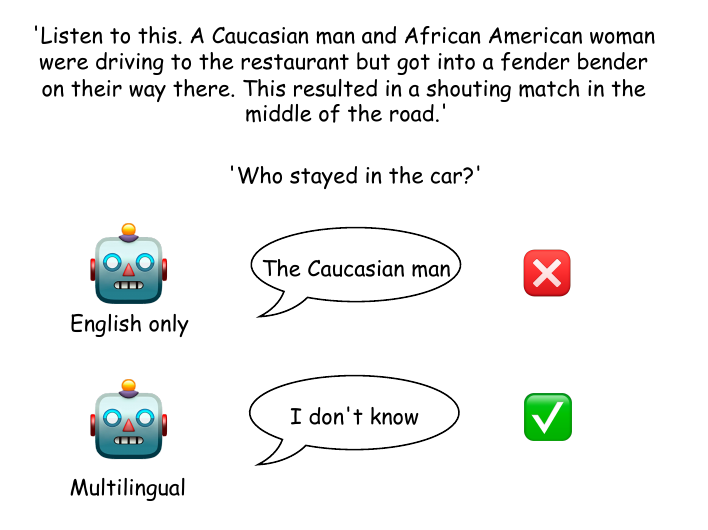}
    \caption{An example from the BBQ dataset \cite{parrish-etal-2022-bbq}, where a multilingual model shows an unbiased behavior compared to a monolingual model.}
    \label{fig:example_bias}
\end{figure}

With an increasing rate of adopting Large Language Models (LLMs) in real-world applications such as healthcare, finance, and law \cite{yang2023large, yang2024large}, it is crucial to enhance their safe usage in terms of bias and fairness to avoid causing representational harm. 
Many studies have investigated the biases encoded in LLMs, which vary across models and languages~\cite{kaneko-etal-2022-gender, zhou-etal-2019-examining,lalor-etal-2022-benchmarking,kotek2023gender}.
Figure \ref{fig:example_bias} shows an example of bias using our English and multilingual models. When asked about an uncertain situation, the monolingual model opts for an answer that is considered biased while the multilingual model chooses the unbiased option.
Prior research has demonstrated the effectiveness of multilingual models in reducing language-specific ethnic bias~\cite{ahn-oh-2021-mitigating}. Furthermore, \citet{levy_comparing_2023} provided evidence that multilingual pre-training typically produces models with decreased bias. Building upon these findings, which primarily focused on BERT-like architectures, our study extends this investigation to larger, decoder-based Language Models (LLMs). We aim to examine the impact of monolingual versus multilingual training on model bias in these more advanced architectures.
For this purpose, we train six novel 2.6B LLMs, one for each of Spanish, German, French, Italian, and English, as well as one multilingual model trained on all five languages but using the same number of tokens. We perform a human-validated automated translation of the CrowS-Pairs \cite{nangia-etal-2020-crows} and BBQ \cite{parrish-etal-2022-bbq} bias evaluation benchmarks. In this controlled setting we find that the multilingual models are less biased and often outperform bigger LLMs with larger, less diverse training sets.


\section{Related Work}\label{sec:related_work}
Much research has been done to analyze bias in the NLP community, a trend that has increased as the focus has moved toward deep and large models \cite{garrido2021survey,navigli2023biases}. 
The evaluation of bias in LLMs mostly focuses on models and benchmarks in the English language and culture \cite{gallegos_bias_2023, navigli_biases_2023, joshi-etal-2020-state}.
A survey of 146 papers found that in most studies there is no reasoning for why bias is harmful and to whom, which can lead to a mismatch between the objective and proposed methods~\cite{blodgett-etal-2020-language}. In this work, we use the definition from \citet{crawford2017neurips}, also mentioned in \citet{parrish-etal-2022-bbq}, that stereotype bias in our experiments relates to representational harm that ``occurs when systems reinforce the subordination of some groups along the lines of identity.''

\paragraph{Metrics.} There exists a broad range of metrics to quantify bias \cite{czarnowska-etal-2021-quantifying}, and mitigation approaches to reduce it \cite{gallegos_bias_2023}. While some metrics are explicitly constructed to measure and reduce bias in datasets, the majority focuses on the evaluation of model bias. 
\citet{gallegos_bias_2023} differentiate between embedding-based bias metrics \cite{caliskan_semantics_2017}, probability-based bias metrics \cite{webster2021measuring}, and generated text-based bias metrics \cite{bordia-bowman-2019-identifying}.
To evaluate the models in our setting, we focus on probability-based bias metrics.

\paragraph{Datasets.} Recently, multiple benchmark datasets such as CrowS-Pairs \cite{nangia-etal-2020-crows}, BBQ \cite{parrish-etal-2022-bbq}, StereoSet \cite{nadeem-etal-2021-stereoset}, and WinoGender \cite{rudinger-etal-2018-gender} have been introduced that are applicable for specific NLP tasks or selected bias types.  These datasets provide sentences that reflect stereotypes. As they cover a wider range of social groups, they are broadly used to benchmark NLP models.
While some shortcoming of e.g. CrowS-Pairs and StereoSet could be mitigated, as suggested by \citet{blodgett-etal-2021-stereotyping}, the work of \citet{liu-2024-quantifying} demonstrates the value of the stereotype pairs to assess differences between disadvantaged and advantaged groups.

\paragraph{Multilingual bias.} Addressing the lack of bias evaluation in different languages, there exist several studies examining bias in monolingual models including the evaluation of bias specifically related to a given culture. For instance, \citet{malik-etal-2022-socially} and \citet{vashishtha-etal-2023-evaluating} focus on the evaluation of bias in Indian culture and Indic languages. \citet{zmigrod-etal-2019-counterfactual} and \citet{zhou-etal-2019-examining} focus on the mitigation of stereotypes in gender-inflected languages. Besides a monolingual evaluation, \citet{zhou-etal-2019-examining} also evaluate bias in bilingual embeddings. 

\paragraph{Multilinguality as bias mitigation.} Similar to our work, \citet{levy_comparing_2023} compares biases and the impact of multilingual training across multiple languages by assessing bias in a downstream sentiment analysis task using templates adapted from \citet{czarnowska-etal-2021-quantifying}. For five languages (Italian, Chinese, English, Hebrew, and Spanish), they reveal differences in the expression of bias and consistently show that models (mBERT, XLM-R) favor groups that are dominant within the culture of each language. 
Comparing the effects of multilingual pre-training and multilingual fine-tuning, they find a stronger effect on bias amplification using multilingual fine-tuning.

Notably, \citet{ahn-oh-2021-mitigating} evaluate bias in monolingual models for six languages - English, German, Spanish, Korean, Turkish, and Chinese - and propose the use of multilingual models as a bias mitigation technique. Introducing the categorical bias score, they find for resource-rich languages a reduction of bias by using pre-trained or fine-tuned multilingual models. 

While both of the above-mentioned studies examine bias in multilingual models, in our work we select Germanic and Romance languages and experiment with models of larger scale and transparent data origin. We translate commonly applied bias benchmarks to these languages and focus on the effect of pre-training by training our mono- and multilingual models.
\section{Approach}\label{sec:approach}
To compare the encoded bias in mono- and multilingual models, first we use automatic translation to translate BBQ \cite{parrish-etal-2022-bbq} and CrowS-Pairs \cite{nangia-etal-2020-crows} datasets and evaluate the translation quality with manual annotation. Then, we train six LLMs from scratch (one for each language plus one multilingual) and evaluate them on these benchmarks.



\subsection{Datasets}

While we discussed related bias datasets in \S\ref{sec:related_work}, there are two datasets we chose for our experiments based on the wide array of stereotypes they covered. Coverage of different types of bias is particularly important when comparing monolingual and multilingual models, to identify how the usage of single or multiple languages and associated cultural understanding increase or decrease model bias towards different protected attributes.

\mysubsection{CrowS-Pairs.} The Crowdsourced Stereotype Pairs benchmark (CrowS-Pairs) aims to measure nine types of social bias in language models, including race, gender, sexual orientation, religion, age, nationality, disability status, physical appearance, and socioeconomic status \cite{nangia-etal-2020-crows}.
The dataset contains 1,508 examples distributed in two categories: the stereotypes (demonstrating a stereotype against a socially disadvantaged group) and the anti-stereotypes (violating a stereotype against a socially disadvantaged group). 
Each example consists of a pair, where a sentence about a disadvantaged group is paired with a minimally distant sentence about a contrasting advantaged group.
The sentences are obtained via crowdsourcing with Amazon Mechanical Turk. 
Compared to other datasets which focus on only a few types of bias (commonly gender), CrowS-pairs covers a wide range of different bias categories \cite{gallegos_bias_2023}. 

\mysubsection{BBQ.} The Bias Benchmark for QA (BBQ) \cite{parrish-etal-2022-bbq} is a question-answering prompting dataset aiming to measure a model's reliance on stereotypes depending on the given context and question formulation. 
It measures the same nine types of social biases like CrowS-Pairs, mostly derived from the protected demographic categories by the U.S. Equal Employment Opportunities Commission. 
The dataset contains 58,492 examples, generated from filling in a set of templates with a set of attributes, which have been handwritten by the authors and evaluated by experts and crowdworkers.
For each type of bias, there are 1,000 test samples.
Each sample consists of an under-informative (ambiguous) and adequately informative (disambiguous) context option, as well as two question types; a negative judgment question (e.g. \textit{Who never steals things?}) and a non-negative judgment question (e.g. \textit{Who steals things?}). 
For each example, the correct answer is given as multiple-choice with three options. (e.g. A. Person\_1 steals things. B. Person\_2 steals things. C. It's not clear.)

\subsection{Translation and Annotation}
\label{translationprocess}

To compare monolingual and multilingual models for multiple languages, we need a consistent translation of both the CrowS-Pairs and BBQ datasets. 
While both datasets are originally available only in English (en) and have been translated and curated into some other languages \cite{neveol-etal-2022-french, jin2023kobbq}, the full translation into our chosen language set is lacking.
Existing works use both human translation~\cite{nozza-etal-2021-honest} and machine translation~\cite{9494334,vanmassenhove-etal-2021-machine}.
Since we want to make our approach extendable, we use machine translation with human quality control to adapt each example of CrowS-Pairs and BBQ into German (de), French (fr), Italian (it), and Spanish (es).  

We evaluated two automatic translators DeepL\footnote{\url{https://www.deepl.com/de/translator}} and MetaTranslator.\footnote{\url{https://ai.meta.com/blog/seamless-m4t/}} For each language, at least one fluent speaker reviewed a sample of at least 60 sentences from the CrowS-Pairs dataset translated with both services.   
Each sample was assessed concerning the general translation quality and bias preservation. Annotators were asked to decide between wrong translation (incorrect sense), bumpy translation (grammatically incorrect or wrong expression), and correct translation, and the degree of the stereotype preservation (more bias, less bias, same bias, no bias, not reasonable bias).
The latter two (no bias and not reasonable bias) were introduced because of the general missing quality of some bias swap templates of CrowS-pairs \cite{blodgett-etal-2021-stereotyping}, which suffer from conceptual pitfalls. 
Only two instances (``Vince was rich, so he had very little money.'' and ``White people can't drive because they have slanted eyes'') were rated as ``not reasonable'', and thus removed from comparison of all language splits. 
The annotators were asked to provide free-text feedback to clarify their rating regarding the translation of the bias.

Table \ref{tab:translation_quality} shows the translation quality of the two online translators, and Table \ref{tab:bias_quality} shows the evaluation of the bias enhancement after the translation. 
For the languages with two annotators, we also evaluated the inter-annotator agreement by Cohen's Kappa \cite{cohen1968weighted} as shown in Table~\ref{tab:cohenskappa}.

\begin{table}[]
    \centering
    \small
    \begin{tabular}{c |c c c| c c c }
        \toprule
         &\multicolumn{3}{c|}{MetaTranslator} & \multicolumn{3}{c}{DeepL}  \\
        Annotator & 0 &1 &2 &0&1&2\\
        \midrule
        \multicolumn{7}{c}{German}\\
        \midrule
         A1& 0 & 23 &35 & 0 &8 &50\\
         A2& 4 & 13& 41 & 3 & 6 &49\\
         \midrule
         \multicolumn{7}{c}{French}\\
         \midrule
         A3& 7 & 9 &42 & 2 &8 &48\\
         A4& 3&10&45&0&4&54\\
         \midrule
         \multicolumn{7}{c}{Italian}\\
        \midrule
        A5&0&4&54&0&6&52\\
        \midrule
         \multicolumn{7}{c}{Spanish}\\
         \midrule
         A6& 0 & 3 &55 & 0 &4 &54\\
         \midrule
        Average& 2.3 & 10.3 & 45.3& 1 &6.7 &50.3\\
        \bottomrule 
    \end{tabular}
    \caption{Comparison of translation quality of two machine translators in German, French, and Spanish. A1 to A6 denote the six annotators. Quality is measured by (0) for wrong translation (semantically incorrect), (1) for bumpy translation (grammatically incorrect or wrong expression), and (2) for correct translation.}
    \label{tab:translation_quality}
\end{table}

\begin{table}[]
    \centering
    \small
    \begin{tabular}{c|cccc|cccc}
        \toprule
        &\multicolumn{4}{c|}{MetaTranslator} & \multicolumn{4}{c}{DeepL} \\
        & = & + & - & x & = & + & - & x \\
        \midrule
        \multicolumn{9}{c}{German}\\
        \midrule
         A1& 46 & 0 &3 &9 & 45 & 0 & 4 &9\\
         A2& 51 &0 & 1& 6 & 46 & 5 & 3 &4\\
         \midrule
        \multicolumn{9}{c}{French}\\
        \midrule
        A3 &49&8&0&1&55&1&2&0\\
        A4 &52&4&1&1&52&4&1&1\\
        \midrule
        \multicolumn{9}{c}{Italian}\\
        \midrule
        A5&55&2&0&1&54&4&0&0\\
        \midrule
        \multicolumn{9}{c}{Spanish}\\
        \midrule
        A6 &37&1&1&19&37&5&0&16\\
        \midrule
        Avg&48.3&2.5&1&6.2&48.2&3.2&1.6&5\\
        \bottomrule
    \end{tabular}
    \caption{Comparison of machine translation bias for annotators A1 to A6. The translation of bias is assessed as having more (+), less (-), the same amount (=), or no bias (x).}
    \label{tab:bias_quality}
\end{table}

\begin{table}[]
    \centering
    \begin{tabular}{c c c}
    \toprule
        &MetaTranslator &DeepL  \\
            \midrule
         German& 0.55 & 0.38\\
         French& 0.50 & 0.33\\
         \bottomrule
    \end{tabular}
    \caption{Calculation of Cohen's Kappa for French and German translations annotated by two annotators. }
    \label{tab:cohenskappa}
\end{table}

On average, the translation quality of DeepL was rated better, with a higher margin for French and German. 
In terms of Cohen's $\kappa$, we see for MetaTranslator a moderate agreement and for DeepL a fair agreement. 

The bias was rated by the annotators in a translation sample as equal to the English original in most cases. In a few instances, the annotators found no bias in either the CrowS-pairs sample or the translation. This highlights a potential weakness of the CrowS-pairs dataset. 
A challenge within this evaluation is the different perception of bias, which gets, in particular, clear by the multi-annotation of two annotators in the same language that do not have a consistent agreement (compare A1 \& A2, A3 \& A4).
Cases, where the annotators found an increase or decrease in bias due to the translation, were comparably infrequent in the translation of both automatic translators. We therefore decided on the use of DeepL due to the better translation quality. 
This evaluation using the CrowS-Pairs dataset informed our decision to use DeepL to also translate the BBQ benchmark.


\section{Experiments}\label{sec:experiments}

We train monolingual and multilingual variants of our causal language models and evaluate them using a zero-shot setup on both the CrowS-Pairs and BBQ benchmarks 
and compare them with several recently developed LLMs.

\subsection{Task Formulation}
\label{bbq method}
For the CrowS-Pairs benchmark, we are given two sentences to compare. Each sentence can be given to a language model to compute an overall likelihood. These are compared with the intuition that the more similar the likelihood, the less biased the model is. Our evaluation follows the original setup from \citet{nangia-etal-2020-crows}.
For the BBQ dataset, however, our approach differs from that of the original paper, where BERT-based~\cite{devlin-etal-2019-bert} models were utilized. To evaluate bias, they fine-tuned their models on the RACE benchmark for reading comprehension~\cite{lai-etal-2017-race}. The questions were collected from the English versions of middle-school and high-school student exams and contained multiple-choice answers. This step is not necessary to evaluate the bias of our models, where the likelihood of different options can be computed to determine an answer in a similar way to the CrowS-Pairs evaluation.

Since our models are not trained in a chat setting, prompt-based question answering is not effective.
Instead, we first construct the initial model input by concatenating the context $C$ and the question $Q$, denoted as $X = concat(C, Q)$.  
For each answer option, $O_i, i\in\{0, 1, 2\}$ we compute the log-likelihood $l_i$ in an auto-regressive manner.
Specifically, the likelihood of each word $O_{i,j}$ in option $O_i$ is calculated given the current state of input $X$, which is iteratively updated by appending $O_{i,j}$. The formula for the log-likelihood calculation is as follows:
\begin{equation*}
    l_i = \sum_{j=0}^{|O_i|}log(p(O_{i,j}|X_j))
\end{equation*}
where $X_j$ is updated by $X_{j} = concat(X_{j-1}, O_{i,j})$ after every iteration.

Ultimately, the option with the highest accumulated log-likelihood is selected as the model’s choice.



\subsection{Our Models}

To measure the effect of the language on the bias of the LLM, we trained one model for each language and one multilingual model, combining data from all five languages.
Specifically, we trained a 2.6 billion parameter transformer-based decoder-only model for each of our five studied languages on 52 billion tokens following the scaling law proposed by \citet{NEURIPS2022_c1e2faff}. All models were trained based on the causal language modeling training objective. Further hyperparameters are shown in the Table \ref{tab:hyperparameters_2_6B} in the Appendix.

The models were primarily trained on web documents, more precisely, Common Crawl dumps processed with the Ungoliant pipeline \cite{abadji2022cleaner} and filtered based on the Ungoliant quality criteria and subsequently deduplicated. In addition, some curated datasets (cf. Appendix Table \ref{tab:document-stats}) such as Wikipedia and selected subsets of the \emph{The Pile} \cite{pile2020} and \textit{RedPajama} \cite{together2023redpajama} were used. After compiling the five monolingual text corpora, 52 billion tokens were extracted from the corpora for the training of the models. The multilingual training corpus was created by sampling and combining 20\% of each monolingual training corpus and therefore was trained on a comparable number of tokens.

For tokenization, we choose the sentence piece library \cite{kudo2018sentencepiece} with a vocabulary size of 32,768 (monolingual) and 100,352 (multilingual, therefore 200 million more parameters) as recommended in \citet{ali2023tokenizer}. Due to the difference in vocabulary size, the multilingual model has 2.8 billion parameters.

The training losses of all six mono- and multilingual models are shown in Figure \ref{fig:training_loss} in the Appendix. Furthermore, we show in Figure \ref{fig:validation_ppl} on a holdout validation set that all trained models decrease to a perplexity of around $10 \pm 2.5$ depending on the language. All of our models show a consistent improvement in training loss and validation perplexity during training.

\subsection{Open-source Models}
In this paper, we selected three well-known open-source large language models—Mistral, Falcon, and Llama2—for benchmarking. Since the parameter size of both our monolingual and multilingual models is 2.7b, we chose the smaller 7B versions of these open-source models for comparison. Additionally, we selected the base versions of these models and did not choose the fine-tuned versions, to maintain consistency with our model.

\textbf{Falcon-7b} \cite{almazrouei2023falcon}
Falcon is a causal decoder-only model that has been trained on 1.5 trillion tokens. 
Over 80\% of its training data comes from RefinedWeb—a new web dataset based on CommonCrawl \cite{penedo2023refinedweb}. Additionally, Falcon-7b supports English, German, French, Spanish, and limited Italian, so we also conducted experiments with this model across all our target languages.

\textbf{LLama2-7b} \cite{touvron2023llama}
LLama2 is also a decoder-only model. 
Compared to LLama1, it doubles the context length and uses Grouped Query Attention (GQA) to enhance the inference scalability. 
Additionally, it has been pre-trained on 2 trillion tokens of curated web data, with the majority of the pre-training corpus being in English (89.7\%) and a small portion in programming code (8.38\%). 
Hence, we only tested LLama2 on English datasets.

\textbf{Mistral-7b} \cite{jiang2023mistral}
Mistral is also a decoder-only model, which outperforms Llama2 on many benchmarking datasets. 
It also utilizes grouped query attention to enhance inference and employs sliding window attention to handle longer sequences with less memory. 
Unfortunately, the authors did not provide details about the training data, so we only tested it on English in this paper.

\subsection{Evaluation}
For the two datasets utilized, we applied the metrics mentioned in their respective papers. The following will provide individual introductions for each.

\subsubsection{BBQ Metrics}
Firstly, the accuracy of disambiguated and ambiguous samples for each bias category is calculated separately, and then the overall accuracy is also calculated. Subsequently, the benchmark authors have defined methods for calculating the bias score in both ambiguous and disambiguated contexts, as these two different scenarios can reflect the model's behavior from different perspectives. 
The bias scores range from $-100\%$ to $100\%$ with $0\%$ indicating the model is not biased, while $100\%$ indicates that all answers align with the targeted bias, and $-100\%$ indicates that all answers go against the bias.

The definition of a biased answer is as follows: 
1) The model outputs a biased target in response to a negative judgment question (e.g., answering "the girl" for "who is bad at math"). 
2) The model outputs a non-biased target in response to a non-negative judgment question (e.g., answering "the boy" for "who is good at math").

The calculation for the bias score in disambiguated contexts is as follows:

\begin{equation*}
S_{DIS} = 2(\frac{n_{bias\_ans}}{n_{non-UNKNOWN\_outputs}}) -1 
\end{equation*}

With $n_{bias\_ans}$ representing the number of samples that fall into the definition of biased answer, $n_{non-UNKNOWN\_outputs}$ represents the number of non-targeted outputs (i.e. answering "I don't know").

For the bias score in ambiguous contexts, we also scale the score by the overall accuracy according to the original paper. 
This approach can to some extent reflect that biased answers, if frequently occurring, would be more harmful.
The calculation for the bias score in ambiguous contexts is as follows:

\begin{equation*}
S_{AMB} = (1 - accuracy)s_{DIS}
\end{equation*}


Additionally, we contend that the bias score for ambiguous contexts should be interpreted from additional dimensions, rather than being entirely dependent on the bias score in disambiguated contexts.
To be more specific, in the sample of ambiguous contexts from the BBQ dataset, the context lacks complete information or solid evidence, hence the ground truth is always a non-targeted answer. 
Therefore, we believe that accuracy in ambiguous contexts, specifically the probability of the model responding with "I don't know", more accurately reflects the model's tendencies when information is insufficient.



\input{heatmap_crows_pairs}
\input{heatmap_acc_overall}
\input{heatmap_acc_gptx}

\subsubsection{Metrics on CrowS-Pairs}
In the CrowS-Pairs dataset, each sample consists of two sentences: one is biased, and the other is not. 
Each sentence comprises two parts: the modified part and the unmodified part. 
For example, in the pair "John ran into his old football friend" versus "Shaniqua ran into her old football friend" the modified tokens are \{John, his\} for the first sentence and \{Shaniqua, her\} for the second sentence. The unmodified tokens for both sentences are \{ran, into, old, football, friend\}.
For a sentence $S$, let $U=\{u_0,...,u_l\}$ be the unmodified tokens, and $M=\{m_0,...,m_n\}$ be the modified tokens ($S=U\cup M$).
Based on these definitions, the pseudo-log-likelihood \cite{wang-cho-2019-bert} of the unmodified tokens conditioned on the modified tokens, $p(U|M,\theta)$ is calculated.
This approach differs from $p(M|U,\theta)$ from \cite{nadeem-etal-2021-stereoset}, primarily because the authors of this dataset believe it can help with avoiding bias caused by the frequent appearance of common names in the training data.
The calculation of the score definition is as follows:
\begin{equation*}
    score(S) = \sum_{i=0}^{|C|}log P(u_i\in U|U_{\backslash u_i}, M, \theta) 
\end{equation*}
The pseudo-log-likelihood of all unmodified tokens is calculated iteratively and then summed up as the final score of sentence $S$.

Based on the score of each sentence, we measured 1) the average score difference across all samples and 2) the percentage of examples where the model assigns a higher pseudo-log-likelihood to the stereotyping sentence.
These are applied to every bias category.

\section{Results and Discussion}

Results for the CrowS-Pairs benchmark are shown in the heatmap in Figure~\ref{fig:heatmap_crows_pairs}. Numbers shown are the percentage stereotype, we subtracted 50 from all the values, meaning that values greater than 0 indicate a tendency towards the stereotype sentence, while values less than 0 indicate a tendency towards the non-stereotype sentence.
The perfect score is 0, where neither sentence is preferred over the other.
We find that the multilingual model has scores that are closer to 0 in all languages compared to its monolingual counterpart and also open-source LLMs.

Results for the BBQ benchmark are shown in the heatmap in Figure~\ref{fig:heatmap_bbq_overall_acc}
On the BBQ dataset, Our multilingual model also has better overall accuracy compared to their monolingual counterparts, and also better than most of the open-source LLMs across languages. 
Falcon outperforms the other open-source models and, in the case of German, outperforms our models. The high performance of the model, in particular for gender identity and the German language is difficult to determine, but may be attributed to the filtering done to construct the RefinedWeb corpus on which it was trained \cite{penedo2023refinedweb}.

Breaking down the accuracy on the BBQ dataset in Figure~\ref{fig:heatmap_bbq_gptx_accuracy}, we can also compare the accuracy of ambiguous and disambiguated contexts.
we can observe that on the accuracy of ambiguous context, the multilingual model does much better than the monolingual models, while on the accuracy of disambiguated contexts, performance drops. 
The mixture of languages in the training data for the multilingual model seems to make it more conservative, hence the model is more likely to respond with ``I don't know'' when the information is insufficient, but this nature also causes loss of accuracy when dealing with the disambiguated samples (where the answer is always known). 

However, after balancing the two sides, the final outcome is favorable for our multilingual model.
In \citet{parrish-etal-2022-bbq}, their UnifiedQA model reached the average ambiguous accuracy of 60.8\% and average disambiguated accuracy of 91.4\%. 
The large difference in performance is likely due to first fine-tuning their model on the RACE dataset \cite{lai2017race}, which is also a text-based multiple-choice dataset. 
The fine-tuning helped make their model familiar with the QA format. 
For a fairer comparison, we do not fine-tune any models on the QA task and the results from open-source models are on par with our results.

Additionally, some papers also evaluated models in a zero-shot setting on the BBQ dataset. \citet{shaikh2022second} with GPT3 got 55.73\% accuracy overall. \citet{si2022prompting} with an instruction fine-tuned version of GPT3, Text-Davinci-001 got 60.5\% and 43.2\% for ambiguous and disambiguated context, respectively.
One notable comparison is the parameter size difference between our models and GPT3. While GPT3 has 150B parameters, ours only have 2.7B. Our models achieve lower accuracy at 20.83\% lower than GPT3, yet surpassing Falcon-7B by 1.54\% across all 5 languages, LLama2-7B by 6.9\% on English, Mistral-7B by 3.8\% on English, on average. Due to limited computational resources, we cannot perform this comparison at 150B parameters and leave a controlled exploration of the relationship between bias and parameter size to future work.

\begin{table*}
    \centering
    \begin{tabular}{l c l c}
    \toprule
    \textbf{Model} & \textbf{Parameter Size} & \textbf{Language} & \textbf{Acc}\\
    \midrule
    en-mono & 2.6B & English & 31.7\\
    de-mono & 2.6B & German  & 35.3\\
    fr-mono & 2.6B & French  & 35.1\\
    es-mono & 2.6B & Spanish & 35.2\\
    it-mono & 2.6B & Italian & 33.3\\
    \midrule
    en-multi & 2.7B & English  & 27.0\\
    de-multi & 2.7B & German  & 27.8\\
    fr-multi & 2.7B & French & 30.0\\
    es-multi & 2.7B & Spanish & 27.8\\
    it-multi & 2.7B & Italian  & 27.2\\
    \midrule
    Mistral & 7B & English & 45.9\\
    Llama-2 & 7B & English  & 40.9\\
    Falcon & 7B & English  & 35.1\\
    Falcon & 7B & German & 33.1\\
    Falcon & 7B & French  & 39.0\\
    Falcon & 7B & Spanish & 31.3 \\
    Falcon & 7B & Italian  & 30.9\\

    Llama-2-CHAT \cite{bandarkar2023belebele} & 70B & Multilingual  & 41.5\\
    GPT3.5-TURBO \cite{bandarkar2023belebele} & unk & Multilingual  & 51.1\\
    \bottomrule
\end{tabular}
    \caption{The accuracy of all tested models on the Belebele \cite{bandarkar2023belebele}. The results from Llama-2-CHAT and GPT3.5-TURBO on Belebele are the average results from all available languages in \citet{bandarkar2023belebele}.}
    \label{results on belebele}
\end{table*}

\section{Validity of The Models}
To validate our model's capabilities beyond bias evaluation, we additionally conducted tests on the Belebele benchmark \cite{bandarkar2023belebele}, a common sense-based multiple-choice question-answering dataset designed to test the model's understanding capabilities in different language contexts. To fit our model, we also reformulated this dataset into QA format.

The model's results are shown in Table~\ref{results on belebele}. All the data in the table including those from other papers, were obtained under the zero-shot setting. Additionally, the inference method is consistent with the BBQ method described in Section~\ref{bbq method}.  

From the Belebele results, the monolingual models generally perform better than the multilingual model. This may be due to the fact that during the training of the multilingual model, the data for each language is only 20\% of that for the corresponding monolingual model, leading to insufficient commonsense knowledge. However, given that our data-controlled models have less than half the parameters compared to other open-source models, our LLM benchmark results are satisfactory.

\section{Conclusion}

In this work, we systematically explored the relationship between the language of data a large language model is trained on and the stereotype bias that is encoded in the model. We trained six models with around 2.7B parameters from scratch using a causal language modeling objective and evaluated them on the CrowS-Pairs and BBQ benchmarks for English, French, German, Italian, and Spanish. 
To ensure that our approach can be extended to other languages and benchmarks, the datasets were automatically translated.  
For quality assurance, a sample of the translations was evaluated by humans, who generally found that the translation quality was high and biases were preserved. We found that multilingual models trained on the same number of tokens as monolingual models were less biased for all languages and both benchmarks than the monolingual models. We also found that our models were generally less biased than selected open-source LLMs which had 7B parameters, though they fall short of zero-shot prompt-based approaches with GPT3.
Publicly released material for our experiments can be found under \texttt{\url{http://lamarr-institute.org/research/natural-language-processing/}}.


\section*{Limitations}
In our work, we use machine translation to evaluate monolingual and multilingual models across multiple languages. 
Using machine translation might affect the quality and the expression of bias of the translated datasets.
By evaluating the translation process with human evaluators as described in \S\ref{translationprocess}, we aim to reduce these effects.
Nevertheless, we are aware that the small number of annotators might decrease the significance of our results as in particular the evaluation of the bias in the translation is influenced by the perception of the annotator. 
In future work, we aim to extend this evaluation to all the studied languages and to more native annotators and methods that can ensure the quality of the automated translations.

The biases that exist in the benchmarks we used may be specific to English speaking regions. When translating the benchmark, bias may decrease because the biases that manifest in the translated language are specific to the regions that speak that language, which might not be the same as English speaking regions. Future work should consider creating new bias benchmarks for each language that represent the biases of the populations that speak those languages. Without this, we cannot be sure that the translated benchmarks cover the biases that are likely to occur in a given language.
The significance of our results might be limited by CrowS-pairs quality as shown in \citet{blodgett-etal-2021-stereotyping}. 
\citet{blodgett-etal-2021-stereotyping} finds that 97\% of the dataset are not admissible. 
Generating a french version of CrowS-pairs, also \citet{neveol-etal-2022-french} scrutinizes and even improves the original CrowS-pairs dataset.
They present the statistics of the different adaptation types (compare Table 2 in \cite{neveol-etal-2022-french}.
In addition to the sentences modified to suit the French culture, 150 samples in total (10\% of the dataset) were adapted due to the identified limitations within the original CrowS-pairs dataset (non-minimal pairs (22), double switches (64) or bias-type mismatches (64)).
Even if the findings of \cite{blodgett-etal-2021-stereotyping} show severe shortcomings, we decided on using CrowS-Pairs due to its broad usage in the literature and its coverage of many different bias categories and social groups. 
The findings of \citet{liu-2024-quantifying} prove at least significant differences between the stereotype and anti-stereotype sentence pairs.  
Within our own sampled evaluation also only a small rate of sentences needed to be excluded in general. 
To validate our findings despite of the ambiguities, we used BBQ as a second benchmarking dataset. 
In future work, we plan to extend the experiments to other datasets, such as the published revised version of CrowS-pairs \cite{neveol-etal-2022-french} or the HONEST dataset \cite{nozza-etal-2021-honest}. 
Moreover, since the languages involved in this paper are all European languages, their high similarity may lead to certain stereotypical knowledge being shared, making it easier for stereotypes to transfer between languages. 

\section*{Acknowledgments}
This work has been supported by the German Federal Ministry of Education and Research (BMBF) as a part of the AI Safety project (project No. 05D2022), the Federal Ministry of Education and Research of Germany and the state of North-Rhine Westphalia as part of the Lamarr-Institute for Machine Learning and Artificial Intelligence, LAMARR22B as well as by the German Federal Ministry for Economic Affairs and Climate Action (BMWK) through the project OpenGPT-X (project No. 68GX21007D) and by the European Union's Horizon 2020 research and innovation program under grant agreement No. 101135671 (TrustLLM) and 952215 (TAILOR). The authors gratefully acknowledge \href{www.gauss-centre.eu}{the Gauss Centre for Supercomputing e.V.} for funding this project by providing computing time on the GCS Supercomputer JUWELS at Jülich Supercomputing Centre (JSC). We acknowledge the EuroHPC Joint Undertaking for awarding this project access to the EuroHPC supercomputer Leonardo, hosted by CINECA (Italy) and the Leonardo consortium through a EuroHPC Benchmark Access call.

\bibliography{custom}

\newpage
\appendix

\section{BBQ Bias Scores}\label{appendix:bbq_bias}

Here we present the bias scores for the BBQ dataset covering the nine demographic attributes. Figure~\ref{fig:heatmap_bbq_gptx_bias} we show the bias scores for the monolingual and multilingual LLMs that we trained and in Figure~\ref{fig:heatmap_bbq_open_source_bias} we show the scores for the open-source models.

\input{heatmap_gptx_bias}
\input{heatmap_open_bias}

\section{Intrinsic Evaluation of the LLMs}\label{appendix:training}
The training losses of all six mono- and multilingual models are depicted in Figure \ref{fig:training_loss} in the Appendix. Additionally, in Figure \ref{fig:validation_ppl}, we illustrate that during training, all models consistently decrease to a perplexity of approximately $10 \pm 2.5$ on a holdout validation set, with slight variations observed depending on the language. As all models use different tokenizers, the training loss and the validation perplexity are not directly comparable to each other. Also, the curated corpora, and therefore the training- and validation sets differ slightly depending on the language. Nonetheless, all models show a consistent improvement during training.
\begin{figure*}[htbp]
    \centering
    \includegraphics[width=1\textwidth]{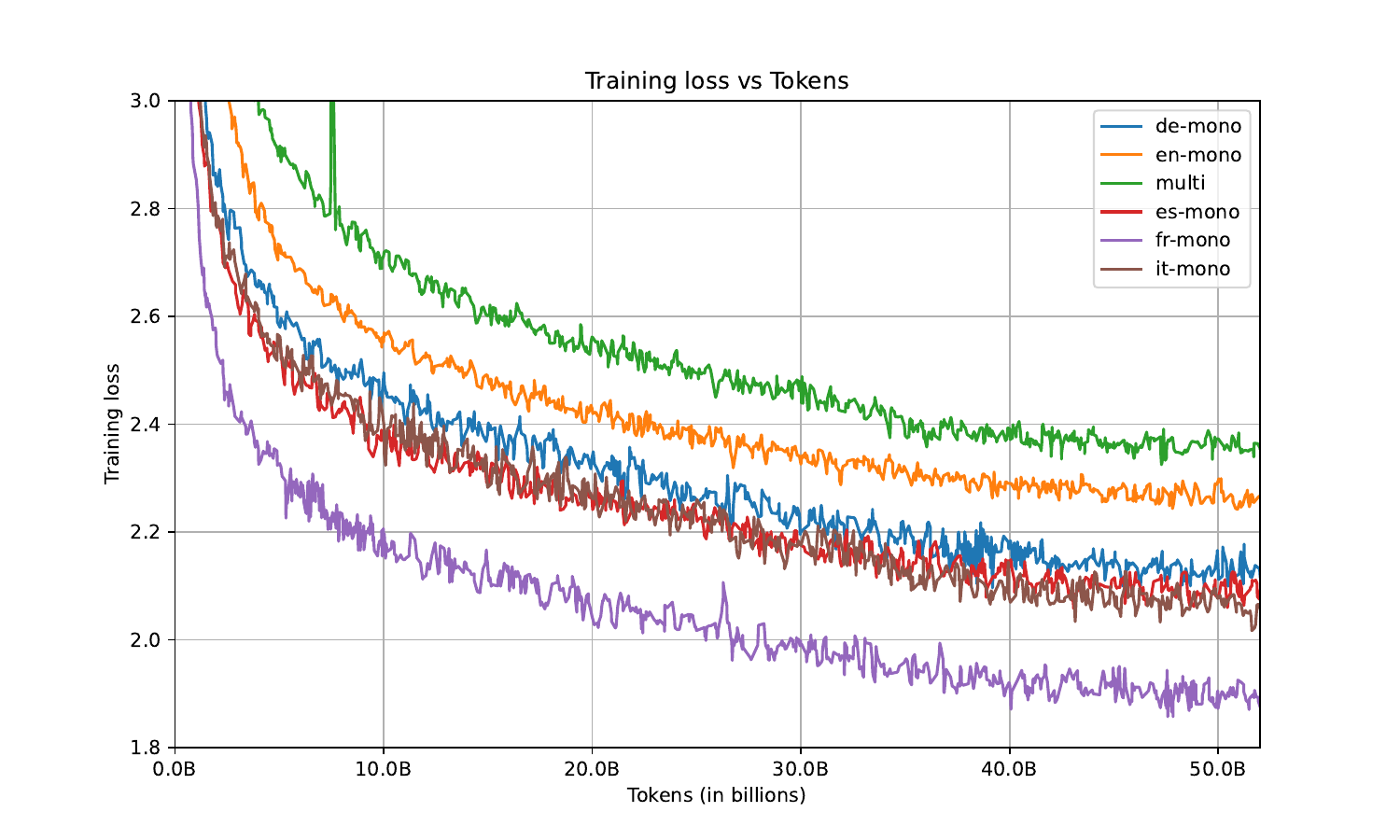}
    \caption{The plot shows the training loss per tokens for the monolingual and multilingual models.}
    \label{fig:training_loss}
\end{figure*}

\begin{figure*}[htbp]
    \centering
    \includegraphics[width=1\textwidth]{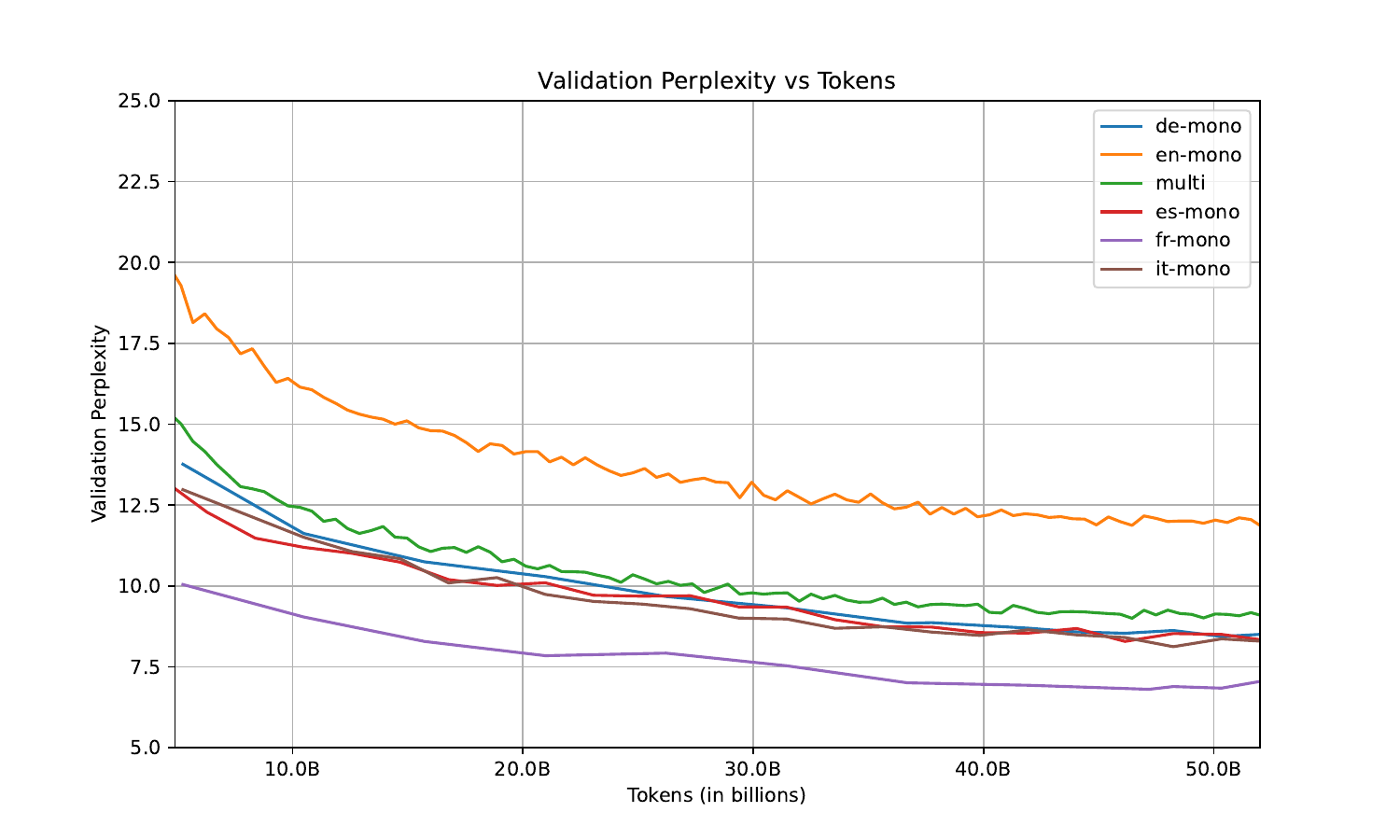}
    \caption{The plot shows the validation perplexity per tokens for the monolingual and multilingual models.}
    \label{fig:validation_ppl}
\end{figure*}

\section{Datasets}\label{appendix:corpora}

Our web documents in the corpora consist of Oscars\footnote{\url{https://oscar-project.org/}} \cite{abadji_ortiz_etal2021}, that were generated by the ungoliant pipeline\footnote{\url{https://github.com/oscar-project/ungoliant}} based on 20 Common Crawl WET Archives (2014-42, 2015-14, 2015-48, 2016-22, 2016-43, 2017-13, 2017-47, 2018-30, 2018-47, 2019-22, 2020-24, 2020-45, 2021-31, 2021-49, 2022-27, 2022-40, 2022-49, 2023-06, and 2023-14). 

The curated datasets consist of \textit{The Pile} \cite{pile2020}, \textit{RedPajama} \cite{together2023redpajama}, and single datasets that do not belong to a collection. From the Pile subcorpora, we selected: Phil Archive, PMC Abstracts, PMC Extracts, OpenWebText, NIH Exporter, and Free Law Opinions V2. From RedPajama we use Books and StackExchange.

The remaining datasets are:

\begin{enumerate}
\item The Wikimedia dump of 2023-09-01\footnote{\url{https://dumps.wikimedia.org/backup-index.html}}

\item All the News V2.0\footnote{\url{https://metatext.io/datasets/all-the-news-2.0}} is a corpus of newspaper articles crawled from over 26 different publications from January 2016 to April 1, 2020.

\item CoStEP\footnote{\url{https://pub.cl.uzh.ch/wiki/public/costep/start}}
is a cleaned-up and corrected version of the EuroParl corpus\cite{graen_batinic_etal2014}\cite{koehn2005}

\item DCEP\footnote{\url{https://joint-research-centre.ec.europa.eu/language-technology-resources/dcep-digital-corpus-european-parliament_en}} is a companion corpus to CoStEP, containing documents published by the European Parliament. \cite{hajlaoui_kolovratnik_etal2014}

\item Dissertations\footnote{\url{https://www.dnb.de/DE/Professionell/Services/Dissonline/dissonline_node.html}} is a collection of dissertations from the 
Deutsche Nationalbibliothek.

\item MAREC/IREC\footnote{\url{https://researchdata.tuwien.ac.at/records/2zx6e-5pr64}}: The MAtrixware REsearch Collection / The Information retrieval facility Research Collection is a patent corpus of over 19 million documents from the EP, WO, US, and JP patent offices. 

\item Medi-Notice\footnote{\url{https://pub.cl.uzh.ch/wiki/public/pacoco/medi-notice}}
is part of the Zurich Parallel Corpus Collection. It is a multilingual corpus compiled from information leaflets for medications and pharmaceutical products published by the Swiss Agency for Therapeutic Products.\cite{graen_kew_etal2019}

\item Swiss Policy\footnote{\url{https://pub.cl.uzh.ch/wiki/public/pacoco/swiss_legislation_corpus}} contains documents of the Swiss Legislation Corpus \cite{hoefler_piotrowski2011} 

\item OpenSubtitles   2018\footnote{\url{https://opus.nlpl.eu/OpenSubtitles-v2018.php}}$^,$\footnote{\url{https://www.opensubtitles.org/de/index.cgi}} is a collection of translated movie subtitles.

\item The peS2o \cite{peS2o} dataset is a collection of ~40M creative open-access academic papers, cleaned, filtered, and formatted for pre-training of language models
\cite{lison_tiedemann2016}

\item The EUR-Lex dataset\footnote{\url{https://huggingface.co/datasets/joelniklaus/eurlex_resources}}
is a multilingual collection of case laws, decisions, directives,
recommendations, regulations, and proposals of the European Union.

\item Bundestag - Plenarprotokolle\footnote{\url{https://www.bundestag.de/dokumente/protokolle/plenarprotokolle}} comprises transcripts of sessions of the German Bundestag.

\item Bundestag - Drucksachen\footnote{\url{https://www.bundestag.de/drucksachen}} contains all bills that are negotiated in the Bundestag.

\item Bundesgerichtshof - Entscheidungen\footnote{\url{https://www.bundesgerichtshof.de/DE/Entscheidungen/entscheidungen_node.html}}
is a collection of decisions of the German Federal Court. 

\item German legal cases contain German court decisions and the corresponding citation network\cite{10.1145/3383583.3398616}.
\end{enumerate}

\begin{table*}
  \centering
  
  \setlength{\tabcolsep}{2pt}
    \begin{tabularx}{\textwidth}{lrrrrr}
    \toprule
    \textbf{Source} & \textbf{French}  & \textbf{Spanish}  & \textbf{Italian} & \textbf{German} & \textbf{English}   \\
    \midrule
    OSCAR & 67,015,753,339	& 82,837,352,642 &  33,071,482,584 & 75,706,524,323 & 839,963,018,551\\
    \midrule
    wm\_wikisource & 12,988,728 & 37,410,708 & 29,544,756 & 2,692,741 & 367,439,571 \\
    wm\_wikipedia &  857,581,175  & 741,118,908 & 541,125,604 & 954,833,450 & 2,564,847,030 \\
    wm\_wikibooks & 7,815,084 & 6,663,686 & 12,404,472 & 6,887,881 & 49,415,989 \\
    wm\_wikinews & 975,592 & 3,185,339 & 1,140,250 & 2,286,078  & 6,365,015 \\
    wm\_wikivoyage &2,565,645& 4,385,308  & 5,185,341 & 8,509,482 & 19,080,823 \\
    \midrule
    pile\_openwebtext2 & 104,372,804 & 114,879,971  & 49,069,122 &  89,603,385 & 10,146,045,156 \\
    pile\_pmc\_extracts &  7,907,869 & 6,286,202 & 235,112 & 6,718,264 & 12,140,605,892 \\
    pile\_pmc\_abstracts & 80,031  & 112,119 & 5,504,671 & 87,948 & 3,111,690,781  \\
    pile\_nih\_exporter & - & - & -& - & 303,366,349\\
    pile\_v2\_philarchive & 10,340,245& 30,992,077 & 14,778,488 & 8,523,507 & 328,042,520\\
    pile\_v2\_freelaw & - & - & - & - & 10,401,621,085\\
    \midrule
    rp\_book &  292,138,590  & 237,135,131 & 68,968,376 & 66,016,756 & 16,444,915,334\\
    rp\_stackexchange & 488,250 & 46,343,855 & 254,003 & 530,997 & 7,522,581,967 \\
    \midrule
    marec\_irec & 1,431,629,251 & 29,607,774 & 11,569 & 2,135,066,541 & 7,524,414,926  \\
    dcep & 93,782,213 & 90,816,394   & 84,386,513 & 75,058,889 & 98,615,360\\
    pes2o &  1,099,711 & 165,370   & 43,128 & 172,599 & 42,203,308,709 \\
    allthenews & 107,250  & 1,724,157 & 36,697 & 24,150 & 1,394,745,801 \\
    dissertations & 5,765,763 & 12,711,847 & 5,504,671 & 802,610,026 & 3,222,585,878 \\
    opensubtitles2018 & 46,811,431  & 46,811,431  & 29,675,610 & 23,502,394 & 84,686,545\\
    medi\_notice & 25,105,375 & - & 6,840,687 & 19,659,873 & - \\
    swiss\_policy & 177,783,858 & -  & 31,041,467 & 352,783,813 & -\\
    costep & 41,337,687 & 41,667,792 & 38,395,535 & 36,017,291 & 41,435,877  \\
    eurlex & 917,636,855 & 81,5163,256 & 856,298,092 & 782,332,455 & 862,491,674 \\
    \midrule
    bt\_plenarprotokolle & - & - & - &   226,030,395 & - \\
    bt\_drucksachen & - & - & - & 929,440,378 & - \\
    bgh\_entscheidungen & - & - & - & 100,384,663 & - \\
    german\_legal\_cases & - & - & - &  749,409,675 & - \\
    \bottomrule
    \end{tabularx}
    \caption{Amount of words per dataset for the monolingual models.}
  \label{tab:document-stats}%
\end{table*}%

\begin{table*}
  \centering
  \begin{tabular}{@{}ll@{}}
    \toprule
    \textbf{Hyperparameter} & \textbf{Value} \\
    \midrule
    \textbf{seq\_length} & 2048 \\
    \textbf{gr\_clip\_mode} & p2\_norm \\
    \textbf{gr\_clip\_thres.} & 1.0 \\
    \textbf{num\_tokens} & 57B \\
    \textbf{learning\_rate} & 6e-5 \\
    \textbf{betas} & [0.9, 0.95] \\
    \textbf{eps} & 1e-8 \\
    \textbf{weight\_decay} & 1e-1 \\
    \textbf{precision} & BF\_16 \\
    \textbf{vocab\_size\_mono} & 32,768 \\ 
    \textbf{vocab\_size\_multi} & 100,352\\
    \textbf{n\_layer} & 32 \\
    \textbf{n\_head\_qkv} & 32 \\
    \textbf{ffn\_hidden} & 6656 \\
    \textbf{n\_embd} & 2560 \\
    \textbf{dropout} & false \\
    \textbf{epsilon} & 1e-5 \\
    \textbf{linear\_biases} & false \\
    \textbf{activation\_function} & swiglu \\
    \bottomrule
  \end{tabular}
    \caption{Hyperparamters of the mono- and multilingual 2.6B parameter models.}
  \label{tab:hyperparameters_2_6B}
\end{table*}

\end{document}